# Speech Emotion Recognition via Contrastive Loss under Siamese Networks

Zheng Lian
National Laboratory of Pattern Recognition, Institute of Automation Chinese Academy of Sciences, School of Artificial Intelligence, University of Chinese Academy of Sciences
lianzheng2016@ia.ac.cn

Ya Li
National Laboratory of Pattern Recognition, Institute of Automation Chinese Academy of Sciences
yli@nlpr.ia.ac.cn

Jianhua Tao
National Laboratory of Pattern Recognition, CAS Center for Excellence in Brain Science and Intelligence Technology, Institute of Automation Chinese Academy of Sciences, School of Artificial Intelligence, University of Chinese Academy of Sciences
jhtao@nlpr.ia.ac.cn

Jian Huang
National Laboratory of Pattern Recognition, Institute of Automation Chinese Academy of Sciences, School of Artificial Intelligence, University of Chinese Academy of Sciences
jian.huang@nlpr.ia.ac.cn

## ABSTRACT

Speech emotion recognition is an important aspect of human-computer interaction. Prior work proposes various end-to-end models to improve the classification performance. However, most of them rely on the cross-entropy loss together with softmax as the supervision component, which does not explicitly encourage discriminative learning of features. In this paper, we introduce the contrastive loss function to encourage intra-class compactness and inter-class separability between learnable features. Furthermore, multiple feature selection methods and pairwise sample selection methods are evaluated. To verify the performance of the proposed system, we conduct experiments on The Interactive Emotional Dyadic Motion Capture (IEMOCAP) database – a common evaluation corpus. Experimental results reveal the advantages of the proposed method, which reaches 62.19% in the weighted accuracy and 63.21% in the unweighted accuracy. It outperforms the baseline system that is optimized without the contrastive loss function with 1.14% and 2.55% in the weighted accuracy and the unweighted accuracy, respectively.

## KEYWORDS

Discrete Emotion Recognition; Convolutional Neural Networks; Siamese Networks; Contrastive Loss; Cross-Entropy Loss

## 1 INTRODUCTION

With the development of artificial intelligence, there is an explosion of interest in realizing more natural human-computer interaction (HMI) systems. The emotion, as an important aspect of HMI, is also attracting increasing attention.

Speech emotion recognition, as an important aspect of emotion recognition, has changed deeply under the influence of deep learning (DL) [1-3]. The traditional method is a multi-step process [4, 5]. Firstly, original signals are divided into overlapping frames and frame-level features are extracted. Secondly, statistic functions, such as mean and maximum, are utilized to obtain utterance-level features. Finally, emotion classifiers are trained to map utterance-level features into emotion labels. However, the multi-step approach has limitations, which roughly consider global information of all frame-level features and ignore temporal dynamics of frames (or segments). Therefore, it may be difficult to extract emotional information from utterances completely.

To release the limitations of multi-step methods, DL is utilized to consider temporal dynamics of frames (or segments). Neumann et al. [6] combined convolutional neural networks (CNNs) with the attention mechanism [7, 8] to consider temporal dynamics of segments. Keren et al. [9] utilized CNNs in combination with recurrent neural networks (RNNs) to extract global information from frame-level features. Huang et al. [10] incorporated the attention mechanism with bidirectional long-short term memory (BLSTM) to extract salient information from each utterance.

However, prior DL-based work relies on the cross-entropy loss together with softmax as the supervision component, which does not explicitly encourage discriminative learning of features. Therefore, we investigate the contrastive loss function that encourage intra-class compactness and inter-class separability between learnable features in this paper. The contrastive loss was first proposed by Chopra et al. [11] for the face verification task, which was associated with siamese networks. The learning process minimized the contrastive loss function, which made the similarity metric to be small for pairs of faces from the same person, and large for pairs from different persons. Then Norouzi et al. [12] extended the contrastive loss, which utilized a flexible form of the triplet ranking loss during training. Salvador et al. [13] utilized the contrastive loss for a novelty task – image-recipe matching. The contrastive loss was optimized with the semantic regularization to map food images and cooking recipes into a joined cross-modal space.

In this paper, we combine the strengths of DL-based methods with the contrastive loss to learn discriminative audio representations. Two loss functions are cooperated in the training stage: the cross-entropy loss and the contrastive loss. Moreover, we discuss the influence of various processing methods, including feature types, signal lengths, testing methods and pairwise sample selection methods. To our best knowledge, it is the first time to optimize with the contrastive loss in speech emotion recognition.

The rest of paper is organized as follows. In Section 2, we describe the proposed system in detail. Experimental setup and results are illustrated in Section 3 and Section 4, respectively. Section 5 concludes the whole paper and discusses future work.

## 2 SYSTEM DESCRIPTION

In this section, the proposed system is illustrated in two aspects: the system architecture and the loss functions. The flowchart of the proposed system is depicted in Fig. 1.

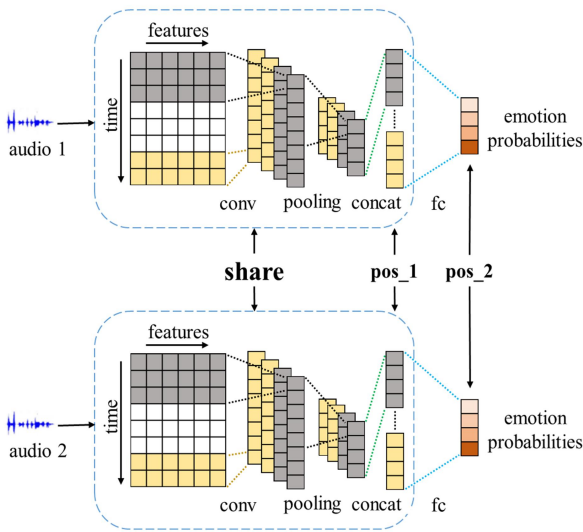

**Figure 1: Flowchart of the proposed system.** There are two branches in the networks that share the same architecture and the same set of weights. (a) The Training Phase: Pairwise audios are sampled from the training dataset. Frame-level features, which are extracted from pairwise inputs, are fed into the CNN architecture to predict emotion probabilities. During training, the contrastive loss function is calculated at "pos_1" or "pos_2". The cross-entropy loss function is calculated at "pos_2". Two loss functions are combined together by a weighting coefficient. The overall loss function is optimized through stochastic gradient descent. (b) The Testing Phase: Testing audios are fed into one branch of the networks. Through the feedforward process, we obtain predictions of testing samples.

### 2.1 System Architecture

There are two branches in the networks, which share the same architecture and the same set of weights. Followed with [6], each branch is a CNN with a convolutional layer and a max pooling layer. The input utterance is divided into $s$ overlapping frames, and each frame is represented as a $d$-dimensional ($d$-D) feature vector. Thus, each utterance can be transformed into a $s \times d$ matrix. Feature selection methods will be illustrated in Sec. 4.

After we extract the $s \times d$ matrix, it is fed into a 1D convolutional layer to extract high-level representations. A batch normalization [14] layer and a nonlinear activation function – ReLU are appended. Then we utilize the max pooling layer to find the salient parts and reduce the feature size. All features are concatenated into a 1D feature vector. To obtain classification results, we pass them through a fully connected layer and a softmax layer in the end.

### 2.2 Loss Functions

Different from previous emotion classifiers [6, 15, 16], we introduce the contrastive loss function for speech emotion recognition. Besides the contrastive loss, we also consider the cross-entropy loss to provide more supervised information in the training stage. Those two loss functions are combined together by a weighting coefficient $\lambda$:

$$L = \lambda L_{con} + (1-\lambda) L_{cro} \qquad (1)$$

where $L_{con}$ and $L_{cro}$ represent the contrastive loss function and the cross-entropy loss function, respectively. And $\lambda$ is the weighting coefficient.

*2.2.1 Contrastive Loss Function.* Let $X_1$ and $X_2$ be a pair of inputs, which represent "*audio 1*" and "*audio 2*" in Fig. 1, respectively. Let $W$ be shared parameters that need to be optimized, and let $G_w(X_1)$ and $G_w(X_2)$ be two points in the low-dimension space that generated by mapping $X_1$ and $X_2$. Then the contrastive loss function can be measured through:

$$D_w(X_1, X_2) = \|G_w(X_1) - G_w(X_2)\| \qquad (2)$$

If $X_1$ and $X_2$ belong to the same category, the contrastive loss function should be small. Otherwise, the loss should be large. In detail, the loss function can be formed as:

$$L(W) = \sum_{i=1}^{N} L\left(W, (Y, X_1, X_2)^i\right) \qquad (3)$$

$$L(W,(Y,X_1,X_2)^i) = \begin{cases} L_G(D_w(X_1,X_2)^i) & Y=1 \\ L_I(D_w(X_1,X_2)^i) & Y=0 \end{cases} \quad (4)$$

where $(Y, X_1, X_2)^i$ is the *i-th* sample, and the dataset is composed with $N$ pairs of audios. $Y$ is the label, which indicates whether two inputs belong to the same category. "$Y=1$" shows that $(X_1, X_2)$ belongs to the same category, which is indicated as positive pairs. Otherwise, it belongs to different categories, which is indicated as negative pairs. $L_G$ is the loss function for positive pairs and $L_I$ is the loss function for negative pairs. $L$ should be designed in such way that the minimization of $L$ will decrease the distance of positive pairs and increase the distance of negative pairs.

In this paper, we test two contrastive loss functions that satisfy the above conditions.

**Contrastive loss type 1:**

$$L(m,(Y,X_1,X_2)^i) = \begin{cases} 1-\cos(X_1^i,X_2^i) & Y=1 \\ \max(0,\cos(X_1^i,X_2^i)-m) & Y=0 \end{cases} \quad (5)$$

$$\cos(X_1^i,X_2^i) = \frac{X_1^i X_2^i}{\|X_1^i\|_2 \|X_2^i\|_2} \quad (6)$$

where $X_1^i$ is the $X_1$ in the *i-th* pairwise sample and $m$ is the margin. Eq. (6) calculates normalized cosine distance between pairwise inputs.

**Contrastive loss type 2:**

$$L(m,(Y,X_1,X_2)^i) = \begin{cases} \|X_1^i - X_2^i\|_2 & Y=1 \\ \max(0,m-\|X_1^i - X_2^i\|_2) & Y=0 \end{cases} \quad (7)$$

where $\|\cdot\|_2$ denotes the $l_2$ norm of a vector and $m$ is the margin.

*2.2.2 Cross-Entropy Loss Function.* This paper incorporates the cross-entropy loss function to consider more supervised information in the training stage. The cross-entropy loss can reduce the difference between predictions and targets.

$$L(p,q) = -\sum_{i=1}^{N} p_i \log q_i \quad (8)$$

where $p$ and $q$ represent normalized possibilities of targets and predictions, respectively. The sum of $p$ (or $q$) is equal to one. $N$ donates the number of categories. $q_i$ represents predictions of the *i-th* category, and $p_i$ represents targets of the *i-th* category.

## 3 EXPERIMENTAL SETUP

The system is tested on The Interactive Emotional Dyadic Motion Capture (IEMOCAP) [17] database – a common evaluation corpus. It contains about 12 hours of audiovisual data, including the video, speech, motion capture of faces and text transcriptions. There are five sessions and each session has two actors. Two actors perform improvisations or scripted scenarios. Sessions are manually segmented into utterances; each utterance is annotated by at least 3 human annotators. There are 10 discrete emotion labels. For this study, we utilize the same category as in [6, 18, 19]: *angry*, *happy*, *sad* and *neutral*. To represent the majority of the emotion categories in the database, *happy* and *excited* are merged into *happy*. For context-independent scenarios, only improvised data are utilized, which are recorded in a pre-defined situation without specific scripts.

To measure the performance in a speaker-independent manner, we utilize the five-folder cross-validation technique. In each folder, four sessions are used as the training set and the validation set. The remaining session is treated as the testing set. The proposed architecture is implemented by Pytorch.

## 4 EVALUATION RESULTS

In this section, three experiments are conducted. In the first experiment, we discuss the influence of different feature selection manners and prediction manners. Since multi-task learning shows its performance in [6, 20, 21], we treat the multi-task learning as the comparison approach in the second experiment. In the last experiment, we show the advantages of adding the contrastive loss during the training phase.

To evaluate the classification performance, we utilize the weighted accuracy and the unweighted accuracy as our evaluation criterion. The weighted accuracy is treated as the primary criterion and the unweighted accuracy is treated as the secondary criterion.

### 4.1 Experiment 1: Feature-level Comparison

In this section, we investigate the influence of different feature selection and prediction methods. To compare the classification performance of various settings, we take one branch in Fig. 1 as the baseline system. The baseline system is optimized without the contrastive loss function by setting the λ in Eq. (1) to be 0.

Since we take fixed-size features as the input, utterances need to be processed into equal length. Longer ones are cut at fixed length and shorter ones are padded with zeros in the end. Such a process is wildly utilized in speech emotion recognition [6, 15]. In this section, we discuss 15 different lengths, ranging from 1 second to 15 seconds. We also discuss two feature sets: 13 Mel-frequency cepstral coefficients (MFCCs) and 26 log-Mel filter-banks features. They approximate the human auditory system's response more closely than the linearly-space frequency used in the normal cepstrum. In the testing phase, we compare two methods: a) average: We evaluate the prediction of the whole utterance by averaging the prediction of all segments [15]. b) crop: We evaluate the prediction of the whole utterance by choosing the prediction of one segment randomly [6].

We utilize the librosa toolkit [22] to extract features from signals. For log-Mel and MFCC features, input signals are segmented into frames by 25ms Hamming windows with the 15ms overlap. For log-Mel features, each frame is passes into Fast Fourier Transformation (FFT) to extract 512-D features. Due to the symmetry of the transformation, only half part is taken. Then we map the powers of the spectrum onto the mel scale over a range from 0 to 6.5 KHz. To extract MFCCs, a discrete cosine transform is taken over log-Mel features. Before training, mean and standard deviation normalization are applied for all features.

Hyper-parameters for the baseline system are listed in Table 1. In the convolutional process, half padding is utilized to make the output size be the same as the input size. We apply 100 kernels for two different filter sizes each, followed with corresponded max pooling layers. To release the impact of the weight initialization, each feature selection method is conducted 20 times.

**Table 1: Details of Hyper-parameters Used for the Baseline System**

| CNN architecture | | |
|---|---|---|
| convolutional layer | No.1 | No.2 |
| # filter outputs | 100 | 100 |
| filter size | 13 | 7 |
| Padding | 6 | 3 |
| stride | 1 | 1 |
| max pooling layer | No.1 | No.2 |
| kernel size | 30 | 30 |
| padding | 0 | 0 |
| Stride | 3 | 3 |

The weighted accuracy and unweighted accuracy of various features and prediction methods are illustrated in Fig. 2. We find that 13 MFCCs are better than 26 log-Mel features, showing that MFCCs are more suitable for emotion classification. Moreover, the weighted accuracy increases with longer inputs in most cases, which verifies that short segments cannot completely convey emotional information of the whole utterance. In the testing phase, "average" is better than "crop" when the fixed signal length is shorter than four seconds, which shows that averaging predictions of all segments can release the influence of the signal length. If the fixed signal length is longer than four seconds, "average" and "crop" have similar results. It shows that emotional information of samples in the IEMOCAP database mainly exists in the first four seconds.

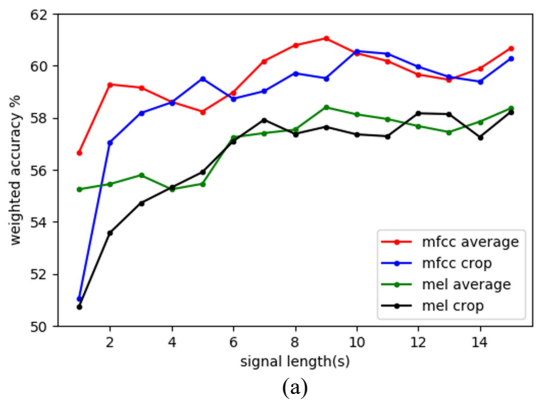

(a)

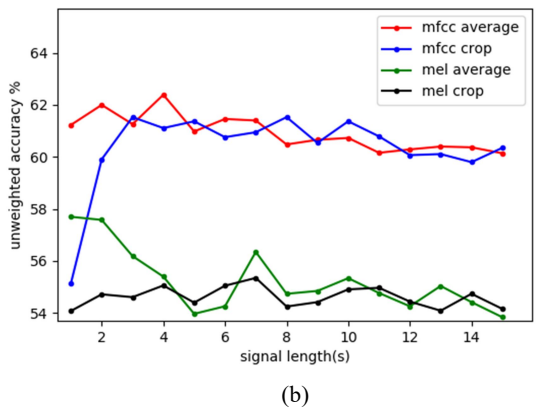

(b)

**Figure 2**: (a): The weighted accuracy (%) of various features and prediction methods. (b): The unweighted accuracy (%) of various features and prediction methods.

In the meantime, we conduct another experiment based on the end-to-end approach, following the architecture in [15]. In this experiment, the log-spectrogram is treated as inputs. However, the results have large variance, where the difference between the highest accuracy and the lowest accuracy is over 10 percentage points. As the end-to-end manner needs more parameters to extract the high-level representations from raw waveforms, it has the overfitting problem for low-resource tasks, which shows the limitation of the end-to-end manner. Therefore, we choose the multi-step approach, where frame-level features are extracted first. Then, we pass them through the CNN networks.

As the highest weighted accuracy in the baseline system is 61.05, 9-second MFCCs is chosen in the following experiments. In the testing phase, we choose "average" to obtain the final predictions.

### 4.2 Experiment 2: Multi-task Learning

Emotions can be represented in two ways: categorical labels (happy, sad, angry and neutral) and dimensional labels (valence, activation and dominance). Furthermore, male and female have different emotion expression manners. In this section, we investigate the influence of other related information through the multi-task learning, which is treated as the comparison experiment.

Following [6, 23], we group valence, activation and dominance into three categories: low: [1,2]; medium: (2,4); high: [4,5]. The gender is treated as a two-way classification task.

Besides the baseline loss in Sec 4.1, we also take into account the cross-entropy loss of other related information such as valence, activation, dominance and gender. To simplify the problem, only one additional task is considered in each time. Those two loss functions are combined together by a weighting coefficient $\lambda$:

$$L = (1-\lambda)L_{baseline} + \lambda L_{addition} \quad (9)$$

where $L_{baseline}$ and $L_{addition}$ represent the baseline loss function and the additional loss function, respectively. The additional loss function refers to the cross-entropy loss of other related information. And $\lambda$ is the weighting coefficient that adjusts the importance of two tasks. As our major task is still the discrete emotion classification problem, $\lambda$ is searched in [0, 0.5]. The highest weighted accuracy of discrete emotions and corresponding $\lambda$ are listed in Table 2.

**Table 2: The Highest Weighted Accuracy (WA) of Discrete Emotions and Corresponding $\lambda$ for Different Tasks**

| Exp. | Additional Tasks | WA (%) | $\lambda$ |
|---|---|---|---|
| 1 | Gender | 61.67 | 0.04 |
| 2 | Valence | 62.00 | 0.08 |
| 3 | Activation | 61.54 | 0.04 |
| 4 | Dominance | **62.20** | 0.04 |
| 5 | None | 61.05 | 0 |

Exp. 5 in Table 2 is the baseline experiment where $\lambda$ is set to be 0, which is also the best result in Experiment 1. Exp. 1~4 in Table 2 cooperate additional tasks through multi-task learning. Through experimental results, we can find that adding additional tasks can gain better performance compared with the baseline system. Dominance can provide the most supplementary information for the target task among all additional tasks. Those combination gains the highest weighted accuracy, 62.20%.

### 4.3 Experiment 3: Impact of Contrastive Loss

In this section, we compare the classification performance of Experiment 1~2 with the proposed architecture.

For the proposed architecture, methods of obtaining pairwise samples are important. In this paper, we test two sampling methods: a) "loader_1": We choose pairwise samples from the training set randomly, and each sample in the training set appears once in "loader_1". b) "loader_2": Firstly, we force each category to have the same number of samples. Then we extract pairwise samples, making that the amount of positive pairs is equal to negative pairs. "loader_2" doesn't cover all training samples.

In this section, we also compare different positions to add the contrastive loss: "pos_1" and "pos_2" in Fig. 1. In the meantime, we discuss the influence of different types of contrastive loss functions: "contrastive loss type 1" and "contrastive loss type 2", which are marked as "loss_1" and "loss_2", respectively.

In the training process, different values of coefficient $\lambda$ in Eq. (1) are tested, ranging from 0.6 to 1.0. The Adam [24] optimizer is utilized to minimize the loss function. The learning rate starts at 1e-4. If the loss value doesn't decrease, the learning rate will be reduced by half. Each experiment is conducted 20 times. Experimental results are shown in Table 3 and Fig. 3~4.

**Table 3: The Highest Unweighted Accuracy (UWA) and the Weighted Accuracy (WA) of Different Configurations**

| Exp. | Pairwise Sampling | Position | Loss | WA (%) | UWA (%) |
|---|---|---|---|---|---|
| 1 | loader_1 | pos_1 | loss_1 | **62.19** | 63.21 |
| 2 | loader_1 | pos_2 | loss_1 | 60.78 | 61.17 |
| 3 | loader_2 | pos_1 | loss_1 | 53.16 | 56.73 |
| 4 | loader_1 | pos_1 | loss_2 | 61.95 | **63.26** |
| 5 | Baseline results | | | 61.05 | 60.66 |
| 6 | The best result in [6] | | | 61.94 | — |

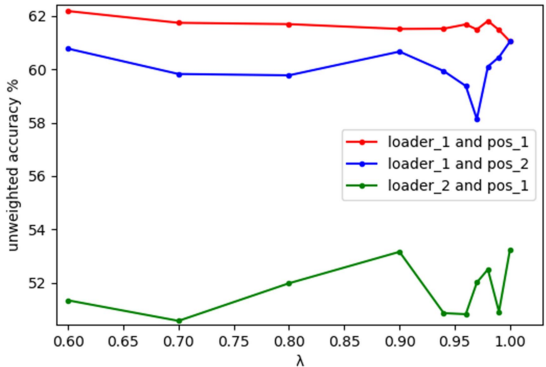

**Figure 3: Comparison of different loaders and positions through the Unweighted Accuracy.**

Through Fig. 3 and Exp. 1~3 in Table 3, we find that "loader_1" is better than "loader_2". We conjecture the reason is that "loader_1" covers more original training samples than "loader_2". Moreover, results show that adding the contrastive loss function at the position "pos_1" is better than "pos_2". The reason may be that the vanishing gradient problem can be released by the early alignment in the training process.

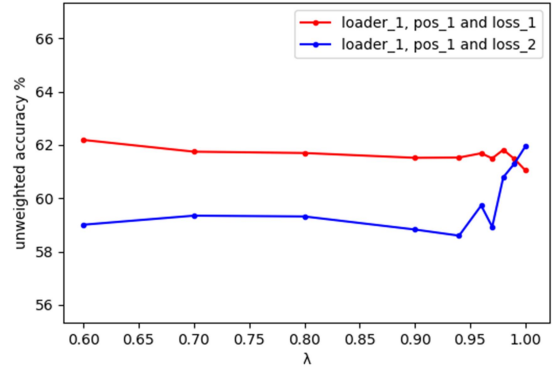

**Figure 4: Comparison of different loss functions through the Unweighted Accuracy.**

Through Fig.4, we can find that the "loss_1" is more stable than "loss_2". Furthermore, "loss_1" can gain better performance compared with "loss_2" in most $\lambda$. However, we can find that their best weighted accuracy has less difference through Exp. 1 and Exp. 4 in Table 3.

If we utilize the same feature set – 13 MFCCs, the same CNN architecture and no additional labels (such as valence and activation), results in [6] still exceed our baseline results according to Exp. 5 and Exp. 6. Our baseline result is between the best result and the worst result reported in [6]. The reason may lie in the difference of weight initialization methods. Although our baseline result is not as good as Exp. 6, the best result we obtained in Exp. 1 is still higher than Exp.6 in the weighted accuracy, which shows the advantages of our proposed method.

Compared with Table 2 and Table 3, we can find that our proposed approach can achieve similar results without considering additional information. It also shows that our method can gain promising results.

Finally, proposed method reaches 62.19% in the weighted accuracy and 63.21% in the unweighted accuracy. It outperforms our baseline system with 1.14% and 2.55% in the weighted accuracy and the unweighted accuracy, respectively.

## 5 CONCLUSIONS

In this paper, we introduce the contrastive loss into speech emotion recognition to extract discriminative audio features. The proposed system has two branches, which shares the same CNN architecture and the same set of weights. The loss functions contain two parts: the cross-entropy loss and contrastive loss. We conduct experiments between different feature selection methods, finding that 13 MFCCs are more suitable for emotion classification problems than 26 log-Mel features. And, the length of input signals also counts. In general, the accuracy decreases with short inputs. Furthermore, we find that sampling methods of the pairwise data are vitally important in the contrastive loss.

We conduct experiments on IEMOCAP. Through experimental results, we find that our proposed approach can achieve similar results without considering additional information, such as valence, activation, dominance and gender. Finally, the proposed system reaches 62.19% in the weighted accuracy and 63.21% in the unweighted accuracy, which outperforms the baseline system with 1.14% and 2.55%, respectively. Those results show the advantages of our proposed method.

In the future, we will extend the system to multimodal recognition systems. The inputs can be replaced by different combinations of the visual modality, the biological modality and the audio modality further. And, we will also combine multi-task learning with the contrastive loss function together to obtain better performance.

## ACKNOWLEDGMENTS

This work is supported by the National Natural Science Foundation of China (NSFC) (No.61425017, No. 61773379), the National Key Research & Development Plan of China (No. 2016YFB1001404) and the Major Program for the National Social Science Fund of China (13&ZD189).

## REFERENCES

[1] A. Stuhlsatz, C. Meyer, F. Eyben, T. Zielke, G. Meier, and B. Schuller, "Deep neural networks for acoustic emotion recognition: raising the benchmarks," in *IEEE international conference on Acoustics, speech and signal processing (ICASSP)*, 2011, pp. 5688-5691: IEEE.

[2] N. E. Cibau, E. M. Albornoz, and H. L. Rufiner, "Speech emotion recognition using a deep autoencoder," *Proceedings of the XV Reunión de Trabajo en Procesamiento de la Información y Control (RPIC 2013), San Carlos de Bariloche*, 2013.

[3] L. Li et al., "Hybrid Deep Neural Network--Hidden Markov Model (DNN-HMM) Based Speech Emotion Recognition," in *Humaine Association Conference on Affective Computing and Intelligent Interaction (ACII)*, 2013, pp. 312-317: IEEE.

[4] C. Vinola and K. Vimaladevi, "A survey on human emotion recognition approaches, databases and applications," *ELCVIA Electronic Letters on Computer Vision and Image Analysis,* vol. 14, no. 2, pp. 24-44, 2015.

[5] P. Chandrasekar, S. Chapaneri, and D. Jayaswal, "Automatic speech emotion recognition: A survey," in *International Conference on Circuits, Systems, Communication and Information Technology Applications (CSCITA)*, 2014, pp. 341-346: IEEE.

[6] M. Neumann and N. T. Vu, "Attentive convolutional neural network based speech emotion recognition: A study on the impact of input features, signal length, and acted speech," *arXiv preprint arXiv:1706.00612,* 2017.

[7] D. Bahdanau, K. Cho, and Y. Bengio, "Neural machine translation by jointly learning to align and translate," *arXiv preprint arXiv:1409.0473,* 2014.

[8] M.-T. Luong, H. Pham, and C. D. Manning, "Effective approaches to attention-based neural machine translation," *arXiv preprint arXiv:1508.04025,* 2015.

[9] G. Keren and B. Schuller, "Convolutional RNN: an enhanced model for extracting features from sequential data," in *International Joint Conference on Neural Networks (IJCNN)*, 2016, pp. 3412-3419: IEEE.

[10] C.-W. Huang and S. S. Narayanan, "Attention Assisted Discovery of Sub-Utterance Structure in Speech Emotion Recognition," in *INTERSPEECH*, 2016, pp. 1387-1391.

[11] S. Chopra, R. Hadsell, and Y. LeCun, "Learning a similarity metric discriminatively, with application to face verification," in *IEEE Computer Society Conference on Computer Vision and Pattern Recognition*, 2005, vol. 1, pp. 539-546: IEEE.

[12] M. Norouzi, D. J. Fleet, and R. R. Salakhutdinov, "Hamming distance metric learning," in *Advances in neural information processing systems*, 2012, pp. 1061-1069.

[13] A. Salvador et al., "Learning cross-modal embeddings for cooking recipes and food images," *Training,* vol. 720, pp. 619,508, 2017.

[14] S. Ioffe and C. Szegedy, "Batch normalization: Accelerating deep network training by reducing internal covariate shift," in *International conference on machine learning*, 2015, pp. 448-456.

[15] A. Satt, S. Rozenberg, and R. Hoory, "Efficient Emotion Recognition from Speech Using Deep Learning on Spectrograms," in *INTERSPEECH*, 2017, pp. 1089-1093.

[16] J. Lee and I. Tashev, "High-level feature representation using recurrent neural network for speech emotion recognition," in *INTERSPEECH*, 2015, pp. 1537-1540.

[17] C. Busso et al., "IEMOCAP: Interactive emotional dyadic motion capture database," *Language resources and evaluation,* vol. 42, no. 4, p. 335, 2008.

[18] R. Xia and Y. Liu, "A multi-task learning framework for emotion recognition using 2D continuous space," *IEEE Transactions on Affective Computing,* vol. 8, no. 1, pp. 3-14, 2017.

[19] S. Ghosh, E. Laksana, L.-P. Morency, and S. Scherer, "Representation Learning for Speech Emotion Recognition," in *INTERSPEECH*, 2016, pp. 3603-3607.

[20] S. Parthasarathy and C. Busso, "Jointly Predicting Arousal, Valence and Dominance with Multi-Task Learning," in *INTERSPEECH*, 2017, pp. 1103-1107.

[21] J. Kim, G. Englebienne, K. P. Truong, and V. Evers, "Towards Speech Emotion Recognition "in the wild" using Aggregated Corpora and Deep Multi-Task Learning," *CoRR,* vol. abs/1708.03920, 2017.

[22] B. McFee et al., "librosa: Audio and music signal analysis in python," in *Proceedings of the 14th python in science conference*, 2015, pp. 18-25.

[23] A. Metallinou, M. Wollmer, A. Katsamanis, F. Eyben, B. Schuller, and S. Narayanan, "Context-Sensitive Learning for Enhanced Audiovisual Emotion Classification," *IEEE Transactions on Affective Computing,* vol. 3, no. 2, pp. 184-198, 2012.

[24] D. P. Kingma and J. Ba, "Adam: A method for stochastic optimization," *arXiv preprint arXiv:1412.6980,* 2014.